\documentclass[11pt]{article}
\usepackage{arxiv}
\usepackage[T1]{fontenc}
\usepackage[utf8]{inputenc}
\usepackage{CJKutf8}
\usepackage{amsmath}
\usepackage{graphicx}
\usepackage{geometry}
\usepackage{booktabs}
\usepackage{float}
\usepackage{hyperref}
\usepackage{setspace}
\usepackage{titlesec}
\usepackage{array}
\usepackage{url}
\usepackage{multicol}

\title{\Large Context-Independent OCR with Multimodal LLMs: Effects of Image Resolution and Visual Complexity}

\author{
 Kotaro Inoue \\
  Independent Researcher\\
  Tokyo, Japan \\
  \texttt{kotaro.inoue.www@gmail.com} \\
}

\begin{document}
\begin{CJK}{UTF8}{min} 

\maketitle

\begin{abstract}
Due to their high versatility in tasks such as image captioning, document analysis, and automated content generation, multimodal Large Language Models (LLMs) have attracted significant attention across various industrial fields. In particular, they have been shown to surpass specialized models in Optical Character Recognition (OCR). Nevertheless, their performance under different image conditions remains insufficiently investigated, and individual character recognition is not guaranteed due to their reliance on contextual cues.
In this work, we examine a context-independent OCR task using single-character images with diverse visual complexities to determine the conditions for accurate recognition. Our findings reveal that multimodal LLMs can match conventional OCR methods at about 300 ppi, yet their performance deteriorates significantly below 150 ppi. Additionally, we observe a very weak correlation between visual complexity and misrecognitions, whereas a conventional OCR-specific model exhibits no correlation. These results suggest that image resolution and visual complexity may play an important role in the reliable application of multimodal LLMs to OCR tasks that require precise character-level accuracy.
\end{abstract}


\section{Introduction}
In recent years, the development of Large Language Models (LLMs) has led to a wide range of applications such as coding assistance, document management, and educational support \cite{hu2024bliva, wu2024next}.
Particularly active is research on multimodal LLMs, which can handle non-verbal information such as images and audio, and their application to fields such as computer vision and creative support is highly anticipated \cite{qu2023layoutllm, huang2024vtimellm}.
Notably, there have been examples in the OCR (Optical Character Recognition) task, where text is generated from images, in which multimodal LLMs demonstrate performance surpassing that of conventional OCR-specific models \cite{shi2023exploring, liu2024textmonkey}.
It is believed that one reason for such high accuracy is the acquisition of an encoder effectively capturing the co-occurrence between images and text through end-to-end training with large-scale paired image-text data.

However, the OCR performance of multimodal LLMs also presents certain challenges. One key issue is that the processes of character recognition and text generation are integrated, making it difficult to evaluate the accuracy of each separately.
For instance, even if some characters in the image are difficult to discern, the system might generate the correct text by relying on contextual clues, leading to a potential overestimation of actual character recognition accuracy.
Although this may have limited impact on text with high contextual dependence, in cases such as random strings of letters or random sequences of numbers, where context does not contribute, the recognition capability of individual characters can be directly tested, and performance degradation may become pronounced \cite{liu2024ocrbench, fu2024ocrbench}.

In this study, we focus on such context-independent OCR tasks and investigate under what image conditions a multimodal LLM can accurately recognize individual characters.

\section{Experimental Methods}

In order to assess the character recognition ability of a multimodal LLM free from contextual supplementation, two key points are considered: the presence of diversity in character shapes, and ensuring that the multimodal LLM is sufficiently trained.
Therefore, this research evaluates OCR accuracy targeting the 2,136 jōyō kanji (commonly used Japanese characters). The experimental procedure is described below.

\subsection{Generation of the Evaluation Kanji Dataset}

The character images for the experiment consist of 2,136 jōyō kanji. These characters were chosen to avoid extremely low-sample conditions for multimodal LLM training. To replicate the condition of characters scanned by a scanner, each character is rendered at a font size of 200 in a 256×256 pixel image. The font used is MS Mincho, which is presumed to be among the most commonly used in Japan.

Evaluating all characters would require substantial time, so we devised a method to sample characters for OCR accuracy evaluation. To create a dataset that incorporates a range of difficulty levels, we quantitatively measure the visual complexity of each character image using fractal dimension and Shannon entropy.
The fractal dimension is a measure that quantifies how completely a geometric object fills space, where higher values indicate more complex and detailed structures. In the context of character recognition, characters with higher fractal dimensions typically have more intricate strokes and details \cite{li2009improved}.
The Shannon entropy, on the other hand, measures the uncertainty or randomness in the distribution of pixel values in an image. Higher entropy values indicate more diverse and complex patterns in the character's visual representation \cite{shannon2001mathematical}.
The fractal dimension is calculated using the box-counting method and is given by the following equation:
\begin{equation}
\displaystyle D = -\lim_{\epsilon \to 0} \frac{\log N(\epsilon)}{\log \epsilon}
\end{equation}
where \(D\) is the fractal dimension, \(\epsilon\) is the box size, and \(N(\epsilon)\) is the number of boxes of size \(\epsilon\) required to cover the image.
In this study, six box sizes of 64, 32, 16, 8, 4, and 2 pixels are used.
The Shannon entropy is calculated based on the distribution of pixel values in the image, expressed by the following equation:
\begin{equation}
\displaystyle H = -\sum_{i=1}^{n} p(i) \log_2 p(i)
\end{equation}
where \(H\) is the Shannon entropy, \(n\) is the range of pixel values, and \(p(i)\) is the probability of pixel value \(i\).
Fractal dimension and Shannon entropy are calculated for each character, and the product of these two values is used as a complexity score. 
A subset of 100 characters is sampled to cover the range of these complexity scores uniformly, thereby creating a dataset that includes characters with various levels of complexity.

The resolution per character depends on both the printed character size and the capabilities of scanners. 
Therefore, we reproduce these conditions by scaling down the images. 
Typically, the font size in an A4 document is about 10 pt, and since 1 pt = 1/72 inch, a scanner with an average performance of 300 ppi corresponds to about 42 pixels. 
In addition, considering that smartphones can serve as an alternative to scanners, we also replicate lower resolutions corresponding to 150, 112.5, and 75 ppi by generating images of 40, 20, 15, and 10 pixels on each side.

\begin{figure}[H]
\centering
\includegraphics[width=0.8\textwidth]{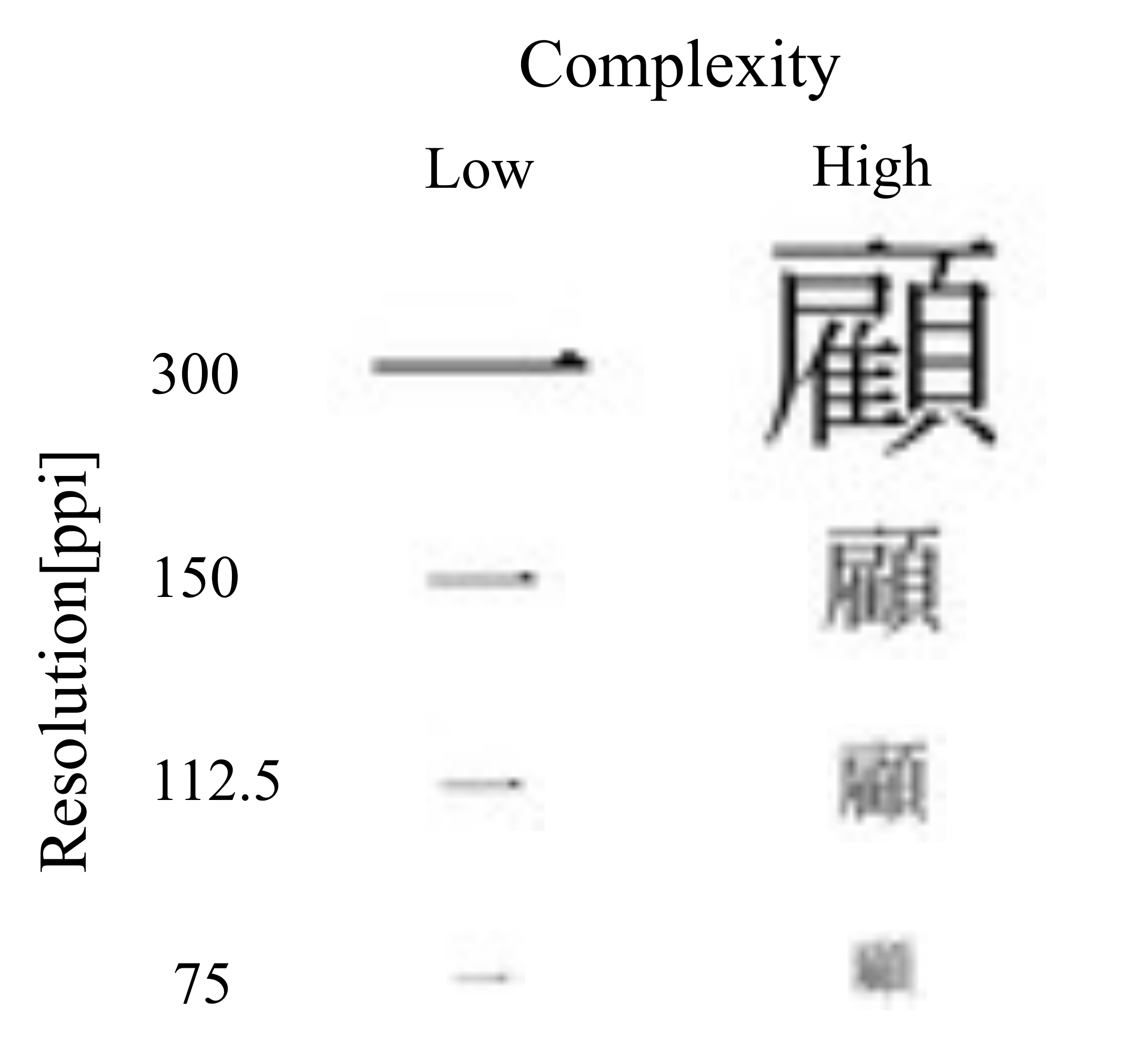}
\caption{Examples of the evaluation kanji dataset}
\label{fig:characters}
\end{figure}

From the 2,136 jōyō kanji, 100 characters are sampled according to their complexity scores, and four types of resolution images are generated for evaluation. In total, 400 character images (100 for each resolution) are created. Figure \ref{fig:characters} shows examples of the evaluation dataset.

\subsection*{OCR Evaluation by Multimodal LLM and OCR Services}

For the generated image samples, OCR processing is performed using two types of multimodal LLMs (GPT-4o and Gemini2.0-Flash) and, as a baseline, the conventional OCR model Azure Computer Vision service. For both multimodal LLMs, we fixed the temperature parameter to 0 to ensure deterministic outputs and used a common prompt "OCR the image, then return the text in JSON format: {"text": str} without code block" across all evaluations to maintain consistency in the experimental conditions. Note that the multimodal LLMs are models as of March 28, 2025, and there is no guarantee that the same results will be reproduced in subsequent years. Also, Azure Computer Vision OCR requires a minimum image resolution of 50 pixels on each side, so padding is applied as needed to expand images to 50×50 pixels.

\section{Experimental Results}
\subsection{Recognition Accuracy for Context-Independent Kanji Characters}

Using the 400 images in the evaluation dataset, OCR processing was performed with the two multimodal LLMs and the existing OCR service.
Comparisons of the OCR results with the correct (ground truth) characters produced an accuracy metric calculated by:
\begin{equation}
\displaystyle \text{Accuracy} = \frac{N_{\text{correct}}}{N_{\text{total}}}
\end{equation}
where \(N_{\text{correct}}\) is the number of correctly recognized characters, and \(N_{\text{total}}\) is the total number of characters.

\begin{figure}[H]
\centering
\includegraphics[width=0.8\textwidth]{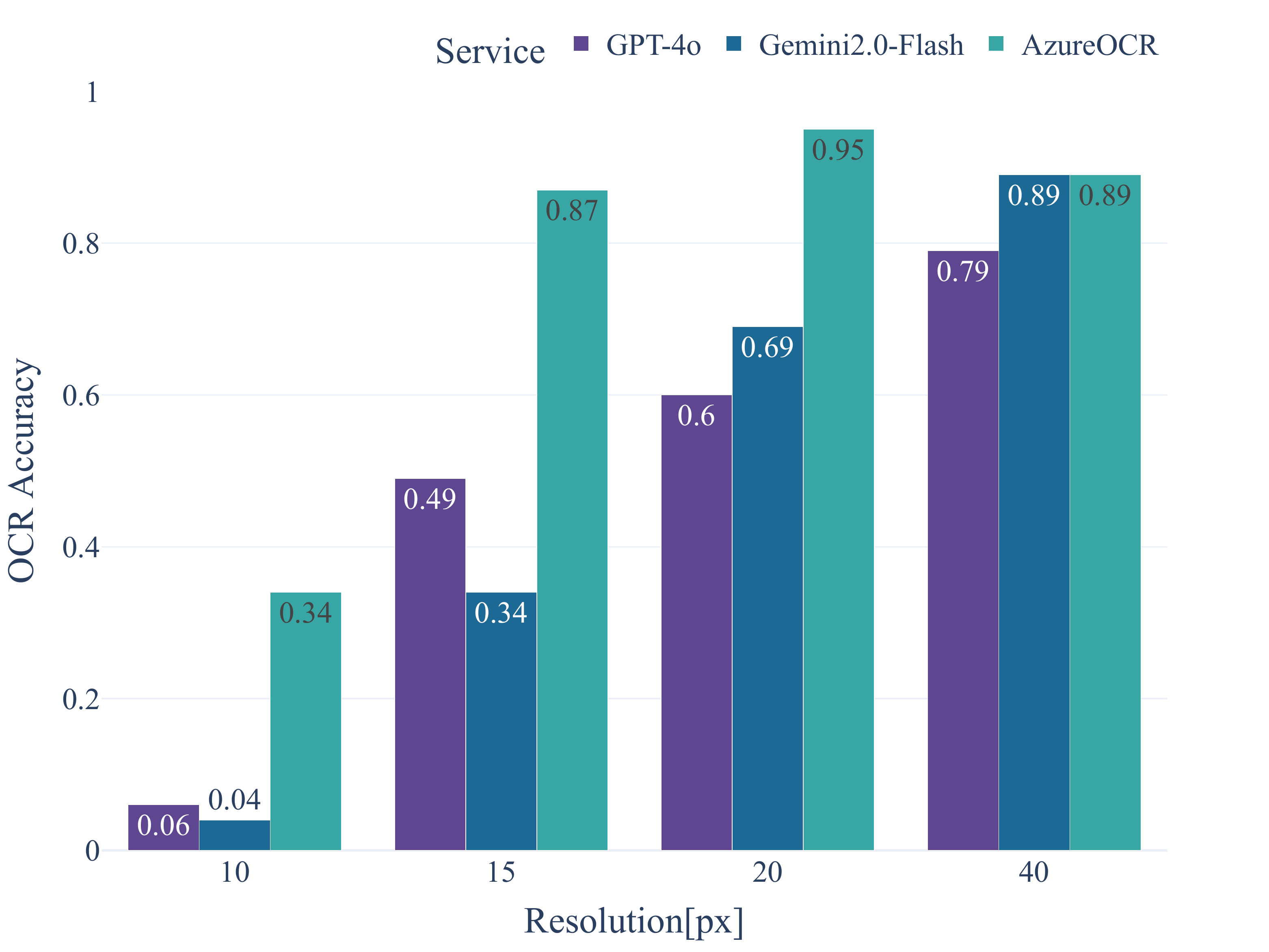}
\caption{Comparison of OCR accuracy at each resolution}
\label{fig:ocr_accuracy}
\end{figure}

Figure \ref{fig:ocr_accuracy} shows the OCR accuracy for each service using this formula.

From Figure \ref{fig:ocr_accuracy}, it can be seen that the multimodal LLMs show almost the same accuracy as existing methods at resolutions equivalent to 300 ppi. However, at resolutions below 150 ppi, the recognition accuracy of multimodal LLMs declines and falls below that of the existing method.

For GPT-4o, which exhibited the highest number of misrecognitions, there were 19 characters consistently misrecognized across all resolutions: "一," "揮," "脅," "緊," "懸," "顧," "栽," "晶," "逝," "戚," "捜," "誕," "墜," "丼," "阜," "紛," "噴," "蔑," "頬." Many are characters with complex structures, but the set also includes characters with simple structures such as "一" and "丼." Most of these misrecognitions involve "一" being read as a hyphen, and "丼" misread as the jōyō kanji "井." While misreading "一" is understandable, the misreading of "丼" would be unlikely for a human, and is thus more difficult to comprehend.

\subsection{Correlation Between Visual Complexity and Misrecognition Trend}

\begin{figure}[H]
\centering
\begin{tabular}{%
  >{\raggedright\arraybackslash}m{0.15\textwidth}%
  >{\centering\arraybackslash}m{0.4\textwidth}%
  >{\centering\arraybackslash}m{0.4\textwidth}}
  & Fractal Dimension & Entropy \\
  GPT-4o &
    \includegraphics[width=0.4\textwidth]{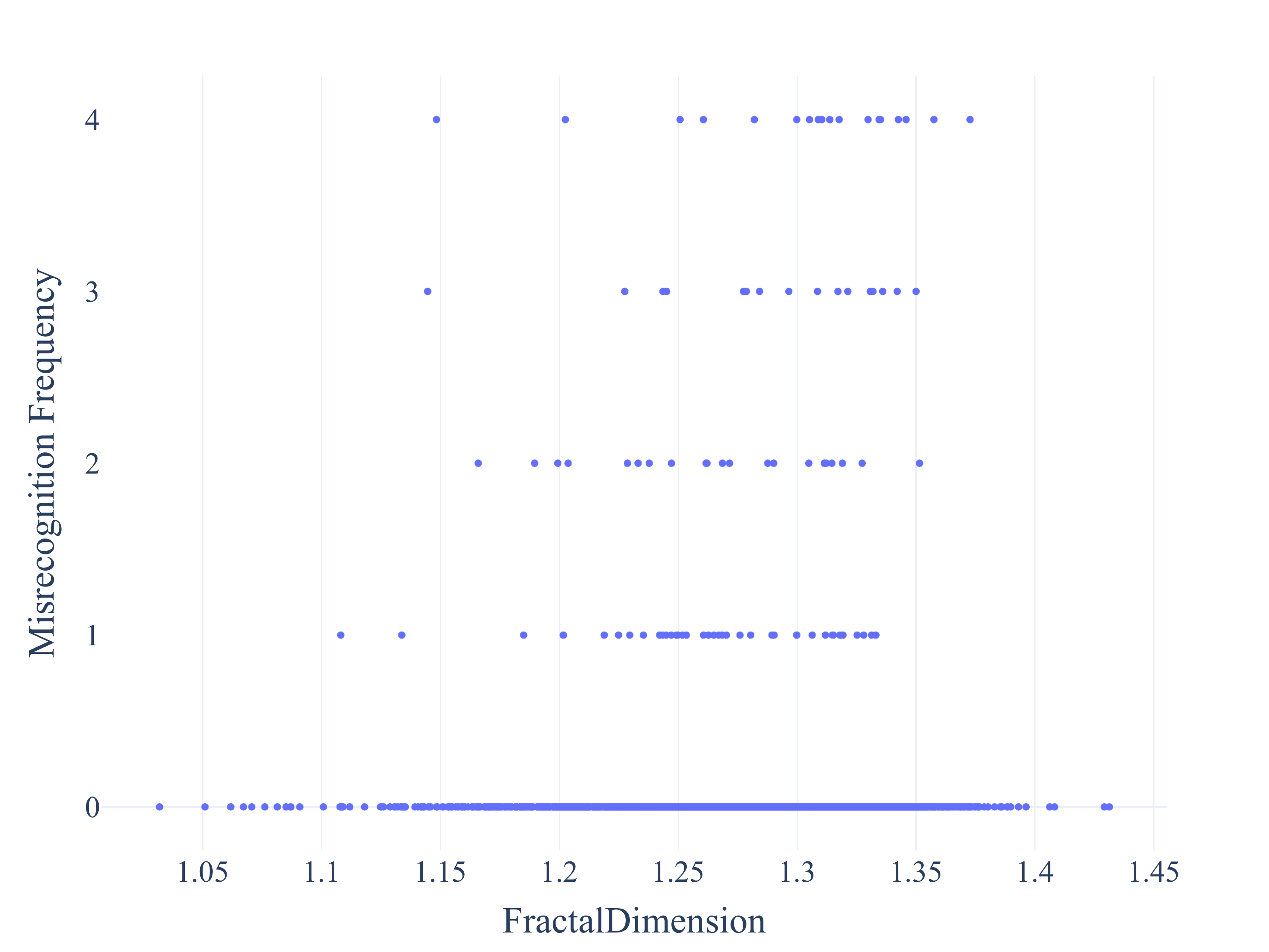} &
    \includegraphics[width=0.4\textwidth]{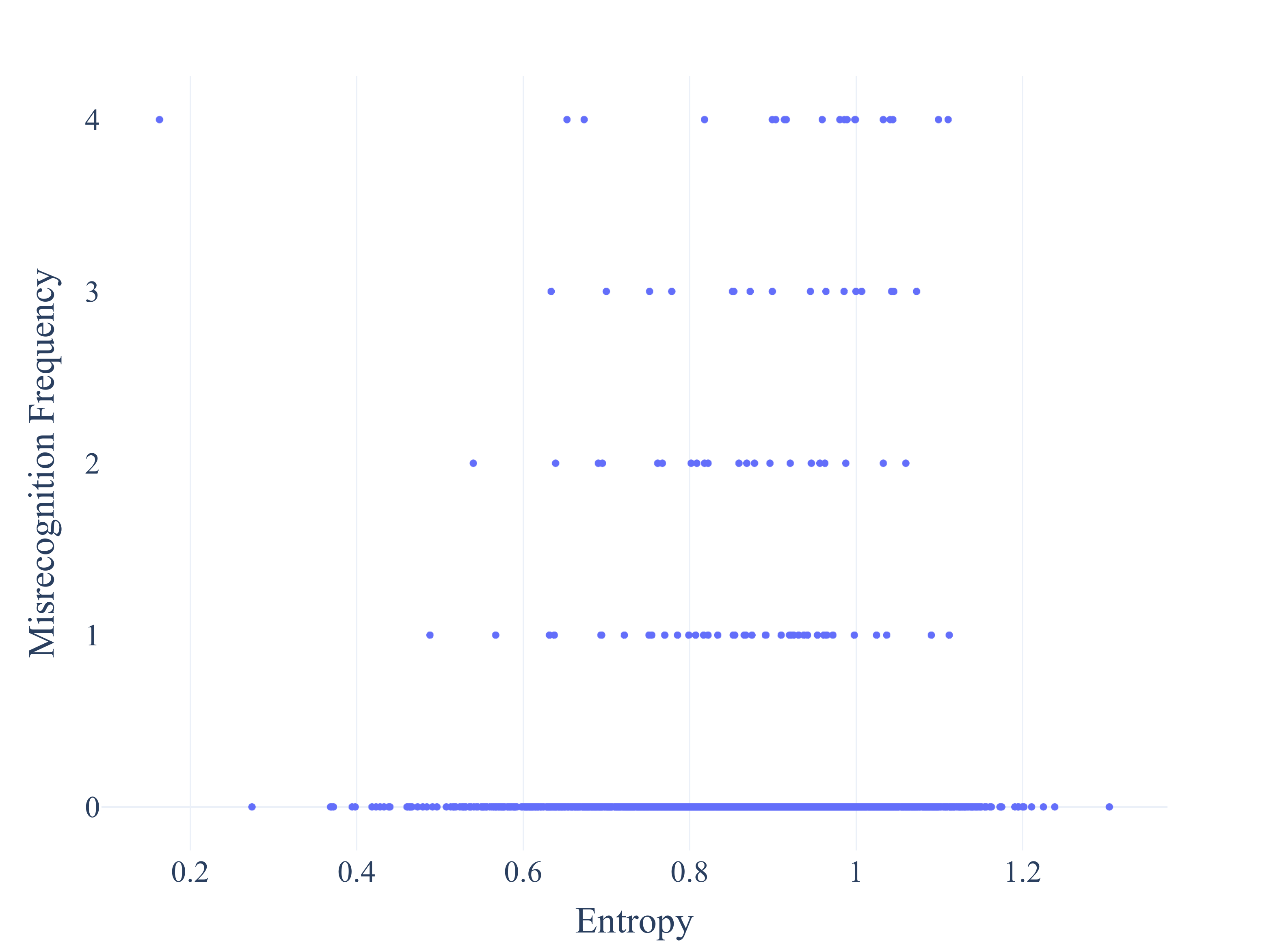} \\
  Gemini2.0-Flash &
    \includegraphics[width=0.4\textwidth]{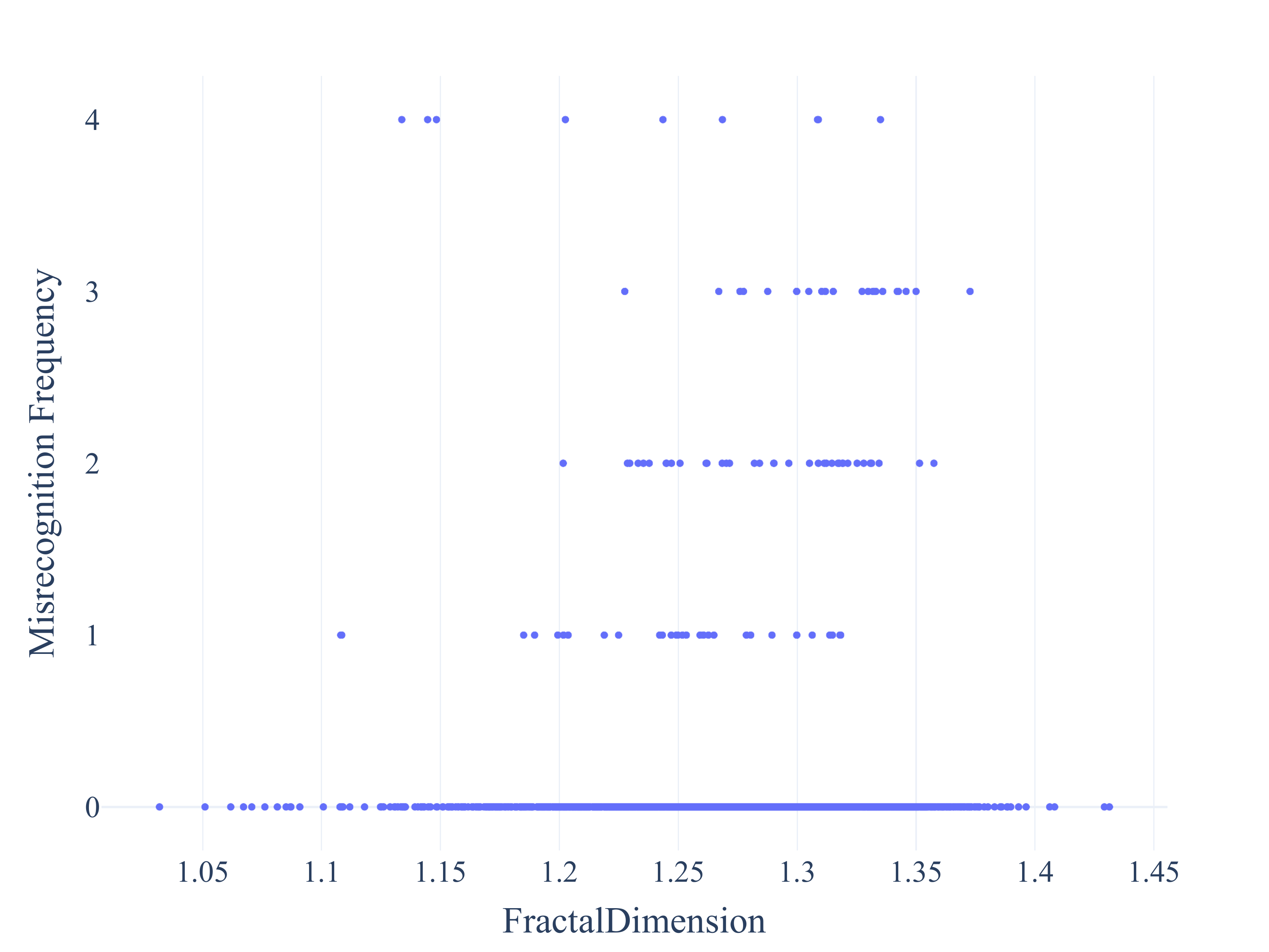} &
    \includegraphics[width=0.4\textwidth]{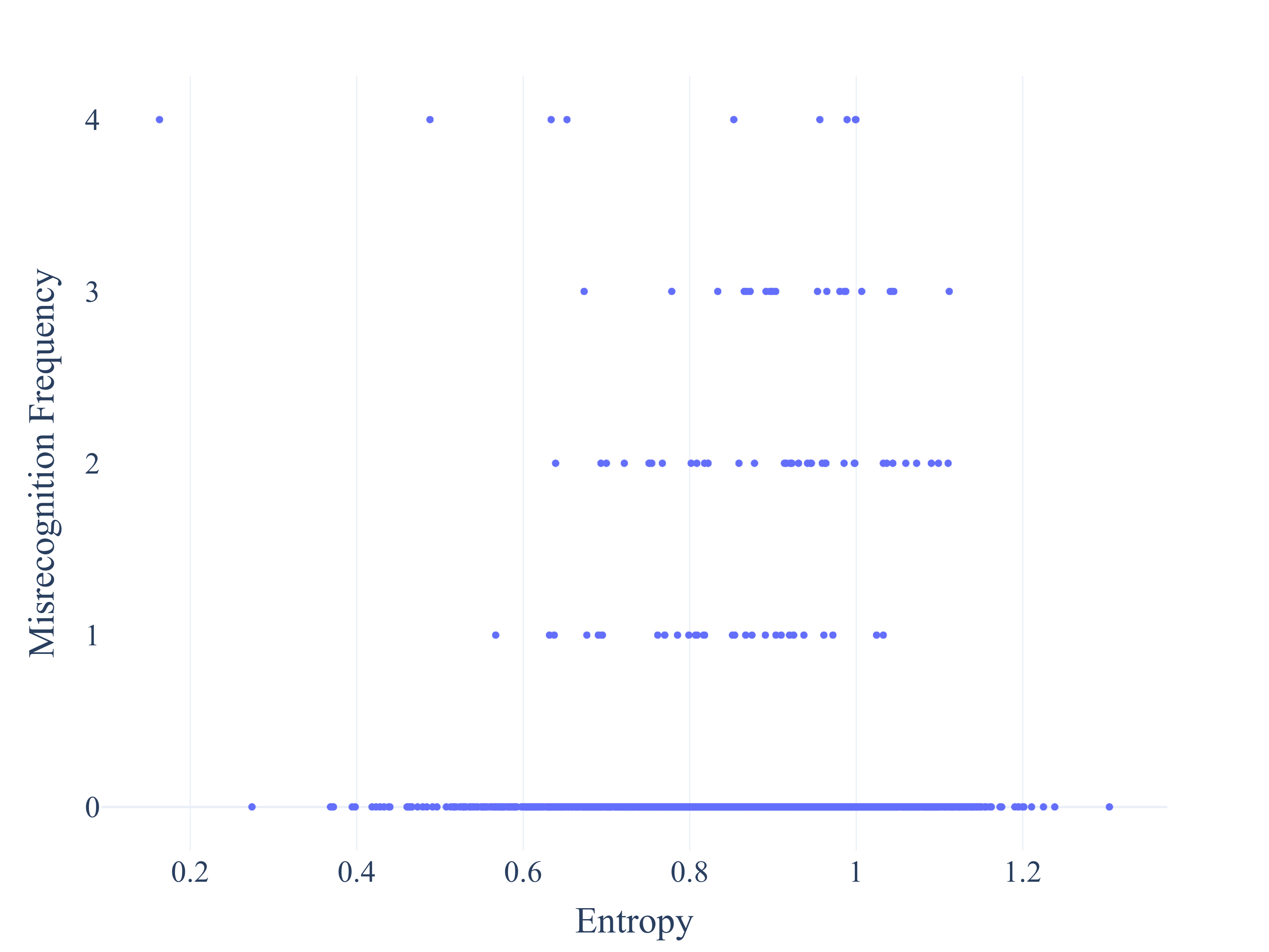} \\
  AzureOCR &
    \includegraphics[width=0.4\textwidth]{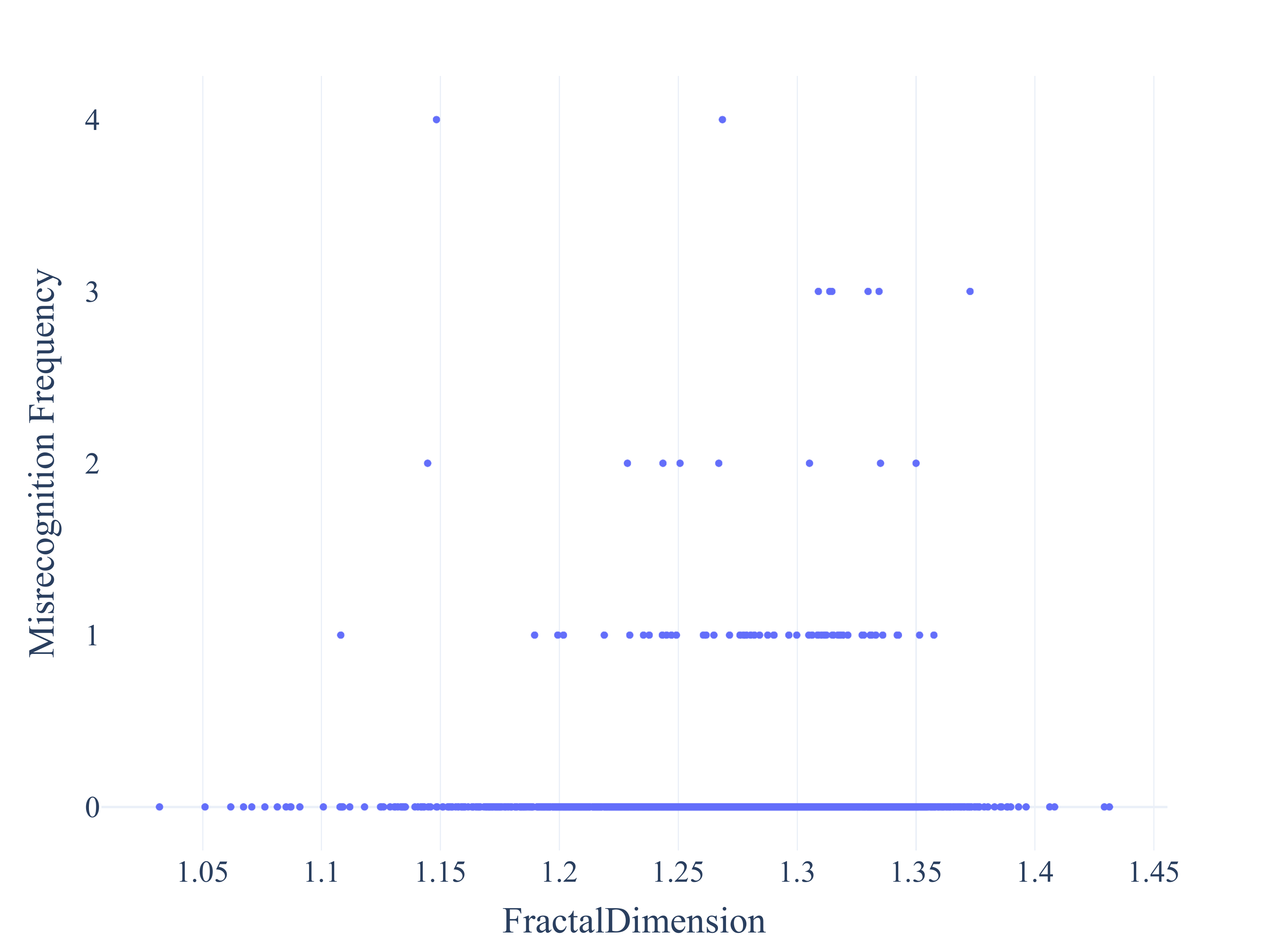} &
    \includegraphics[width=0.4\textwidth]{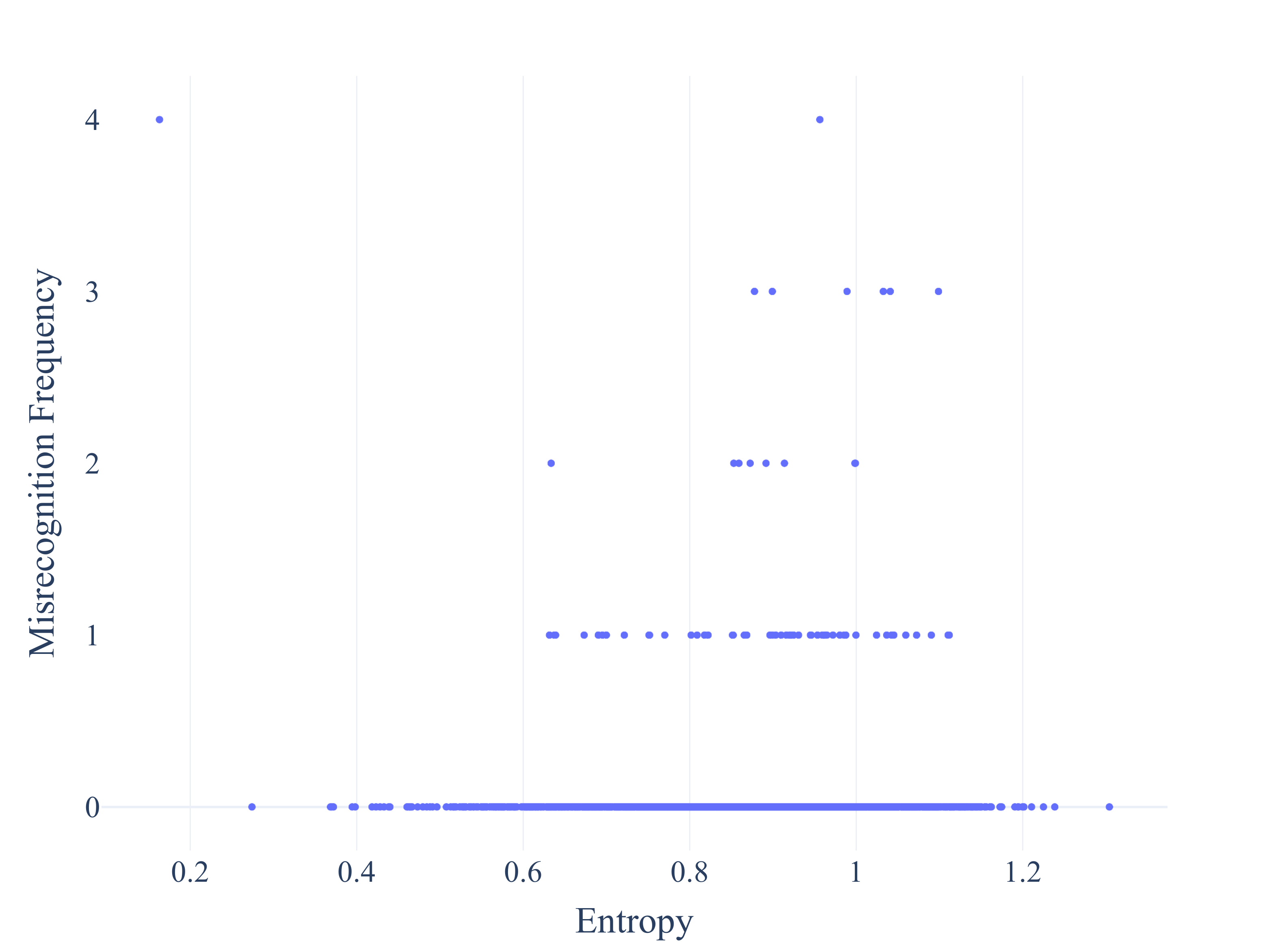} \\
\end{tabular}
\caption{Relationship between visual complexity and misrecognition frequency for each model}
\label{fig:scatter_plots}
\end{figure}

The OCR results from the multimodal LLMs suggest that the complexity of a character might influence its misrecognition frequency. In order to verify this, we examined the misrecognitions across characters at different resolutions and plotted the total number of misrecognitions for each character against their fractal dimension or entropy, as shown in Figure \ref{fig:scatter_plots}.

From Figure \ref{fig:scatter_plots}, we see a gradual increase in the number of misrecognized characters as fractal dimension and entropy increase. Table \ref{tab:correlation} shows the correlation coefficients between the number of misrecognitions (for characters misrecognized one or more times) and fractal dimension or entropy.

\begin{table}[H]
\caption{Correlation between misrecognition frequency and visual complexity for each model}
\centering
\begin{tabular}{lcc}
\toprule
Model & Fractal Dimension Correlation & Entropy Correlation \\
\midrule
GPT-4o & 0.281 & 0.150 \\
Gemini2.0-Flash & 0.172 & 0.037 \\
AzureOCR & -0.031 & -0.116 \\
\bottomrule
\end{tabular}
\label{tab:correlation}
\end{table}

From Table \ref{tab:correlation}, GPT-4o and Gemini (both multimodal LLMs) show a very weak correlation (0.17–0.28) with fractal dimension. On the other hand, no clear correlation was observed between AzureOCR's misrecognition frequency and fractal dimension.

\section{Discussion}

Our results demonstrate that, in the context-independent OCR task of individual characters, the recognition accuracy of multimodal LLMs is significantly influenced by the input image resolution, which in turn depends on the performance of the document scanner. The performance degradation observed at lower resolutions can be attributed not only to the loss of image details but also to potential limitations in the multimodal LLM's visual encoder in processing fine character structures.

Specifically, multimodal LLMs rely on the correspondence between images and text acquired through large-scale paired training of natural images and common texts. For extremely constrained, context-independent data such as single characters, the extraction of relevant visual features may be insufficient. This limitation becomes particularly apparent when processing complex characters at lower resolutions.

From the analysis of misrecognized characters by multimodal LLMs, we observed a very weak correlation between fractal dimension of characters and the misrecognition frequency in GPT-4o. This suggests that more visually complex characters require higher resolution but that the model's capacity to identify structural details may be limited. However, the correlation is weaker in Gemini, indicating possible differences in visual encoder performance or preprocessing between models.

These findings highlight that, when applying multimodal LLMs to OCR tasks in practical scenarios, it is critically important to consider input image quality, particularly resolution. Given the ongoing expansion of applications for multimodal LLMs, our results suggest that in OCR use cases demanding high accuracy without guess-based supplementation (e.g., passwords, ID numbers, monetary amounts, or parts of addresses), specialized preprocessing or the use of a dedicated character-recognition sub-model may become necessary.

\section{Conclusion}
In this study, we evaluated the accuracy of character recognition in multimodal Large Language Models (LLMs) under a context-independent OCR task, using Japanese jōyō kanji that exhibit diverse complexity in their shapes. We analyzed the impact of image resolution and visual complexity of characters on recognition accuracy.
The experimental results reveal that multimodal LLMs achieve accuracy comparable to existing OCR methods at resolutions equivalent to about 300 ppi but show a rapid decline in performance below about 150 ppi.
Furthermore, while there is a very weak correlation between the fractal dimension of characters and the frequency of misrecognitions for multimodal LLMs, no such correlation is observed in conventional OCR models.
These results imply that achieving reliable character-level accuracy in OCR with multimodal LLMs may depend on factors such as image resolution and the intricacy of character forms.

Future work needs to distinguish whether this decrease in OCR performance is attributable to the inherent difficulty of the characters (i.e., visual complexity) or to potential limitations in the visual encoder or preprocessing within multimodal LLMs. More detailed investigations, such as examining open-source implementations of multimodal LLMs, are required to clarify these points.

\bibliographystyle{unsrt}
\bibliography{references}

\begin{thebibliography}{10}

\bibitem{hu2024bliva}
Wenbo Hu, Yifan Xu, Yi~Li, Weiyue Li, Zeyuan Chen, and Zhuowen Tu.
\newblock Bliva: A simple multimodal llm for better handling of text-rich
  visual questions.
\newblock In {\em Proceedings of the AAAI Conference on Artificial
  Intelligence}, volume~38, pages 2256--2264, 2024.

\bibitem{wu2024next}
Shengqiong Wu, Hao Fei, Leigang Qu, Wei Ji, and Tat-Seng Chua.
\newblock Next-gpt: Any-to-any multimodal llm.
\newblock In {\em Forty-first International Conference on Machine Learning},
  2024.

\bibitem{qu2023layoutllm}
Leigang Qu, Shengqiong Wu, Hao Fei, Liqiang Nie, and Tat-Seng Chua.
\newblock Layoutllm-t2i: Eliciting layout guidance from llm for text-to-image
  generation.
\newblock In {\em Proceedings of the 31st ACM International Conference on
  Multimedia}, pages 643--654, 2023.

\bibitem{huang2024vtimellm}
Bin Huang, Xin Wang, Hong Chen, Zihan Song, and Wenwu Zhu.
\newblock Vtimellm: Empower llm to grasp video moments.
\newblock In {\em Proceedings of the IEEE/CVF Conference on Computer Vision and
  Pattern Recognition}, pages 14271--14280, 2024.

\bibitem{shi2023exploring}
Yongxin Shi, Dezhi Peng, Wenhui Liao, Zening Lin, Xinhong Chen, Chongyu Liu,
  Yuyi Zhang, and Lianwen Jin.
\newblock Exploring ocr capabilities of gpt-4v (ision): A quantitative and
  in-depth evaluation.
\newblock {\em arXiv preprint arXiv:2310.16809}, 2023.

\bibitem{liu2024textmonkey}
Yuliang Liu, Biao Yang, Qiang Liu, Zhang Li, Zhiyin Ma, Shuo Zhang, and Xiang
  Bai.
\newblock Textmonkey: An ocr-free large multimodal model for understanding
  document.
\newblock {\em arXiv preprint arXiv:2403.04473}, 2024.

\bibitem{liu2024ocrbench}
Yuliang Liu, Zhang Li, Mingxin Huang, Biao Yang, Wenwen Yu, Chunyuan Li,
  Xu-Cheng Yin, Cheng-Lin Liu, Lianwen Jin, and Xiang Bai.
\newblock Ocrbench: on the hidden mystery of ocr in large multimodal models.
\newblock {\em Science China Information Sciences}, 67(12):220102, 2024.

\bibitem{fu2024ocrbench}
Ling Fu, Biao Yang, Zhebin Kuang, Jiajun Song, Yuzhe Li, Linghao Zhu, Qidi Luo,
  Xinyu Wang, Hao Lu, Mingxin Huang, et~al.
\newblock Ocrbench v2: An improved benchmark for evaluating large multimodal
  models on visual text localization and reasoning.
\newblock {\em arXiv preprint arXiv:2501.00321}, 2024.

\bibitem{li2009improved}
Jian Li, Qian Du, and Caixin Sun.
\newblock An improved box-counting method for image fractal dimension
  estimation.
\newblock {\em Pattern recognition}, 42(11):2460--2469, 2009.

\bibitem{shannon2001mathematical}
Claude~Elwood Shannon.
\newblock A mathematical theory of communication.
\newblock {\em ACM SIGMOBILE mobile computing and communications review},
  5(1):3--55, 2001.

\end{thebibliography}

\newpage
\section*{Appendix}
\subsection*{Evaluation Kanji Dataset}

\begin{figure}[H]
\centering
\begin{tabular}{cc}
\includegraphics[width=0.45\textwidth]{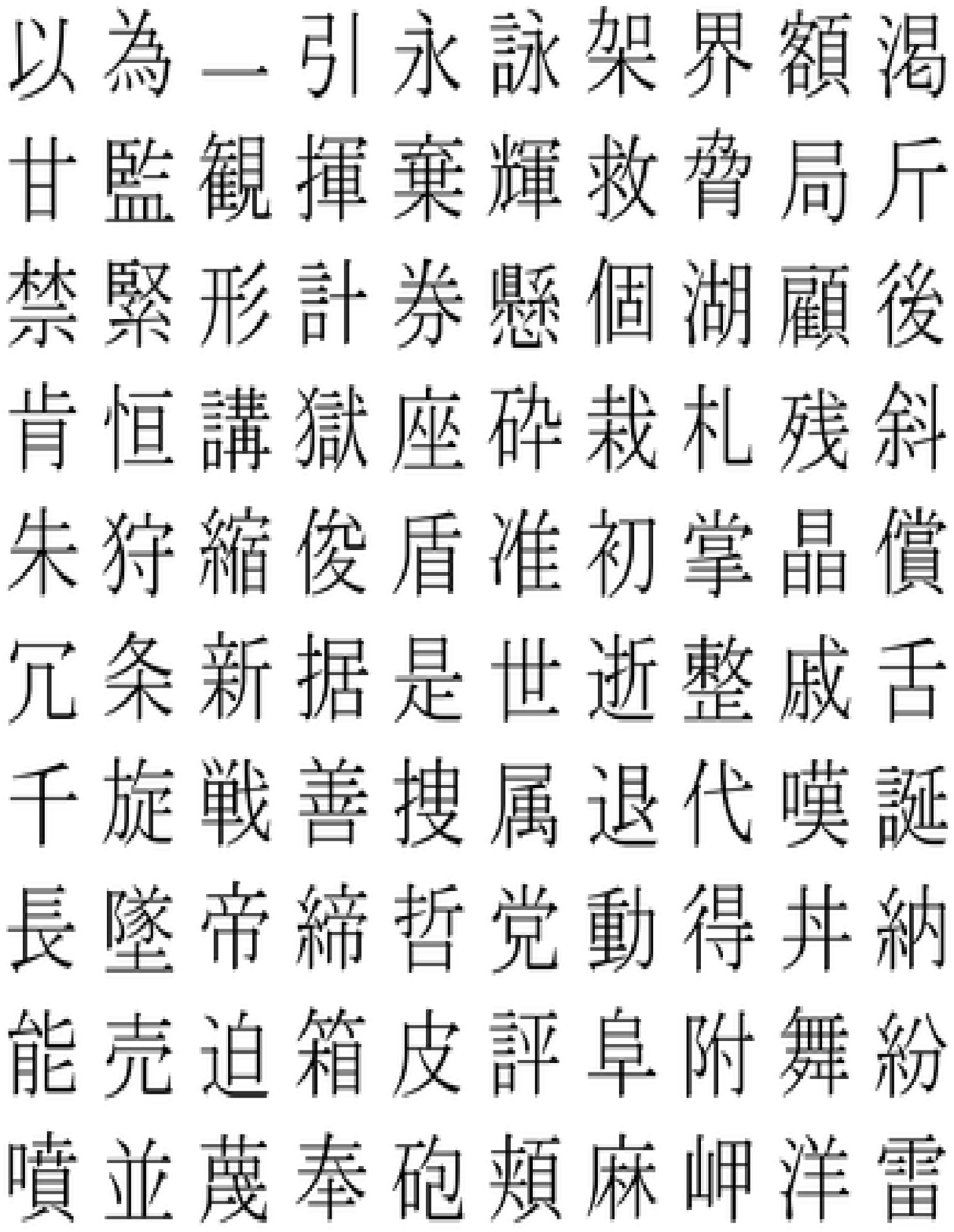} &
\includegraphics[width=0.45\textwidth]{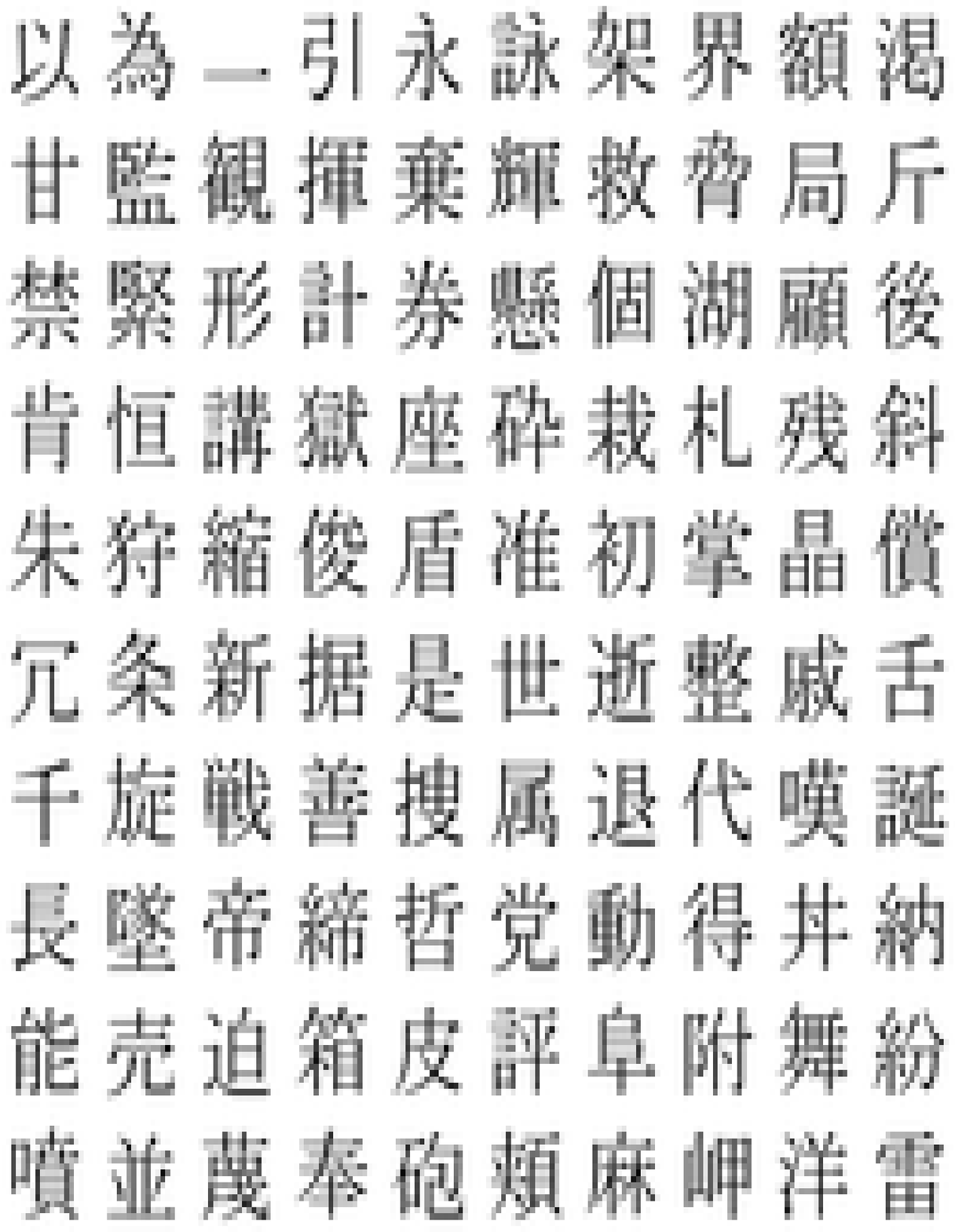} \\
300 ppi & 150 ppi \\
\includegraphics[width=0.45\textwidth]{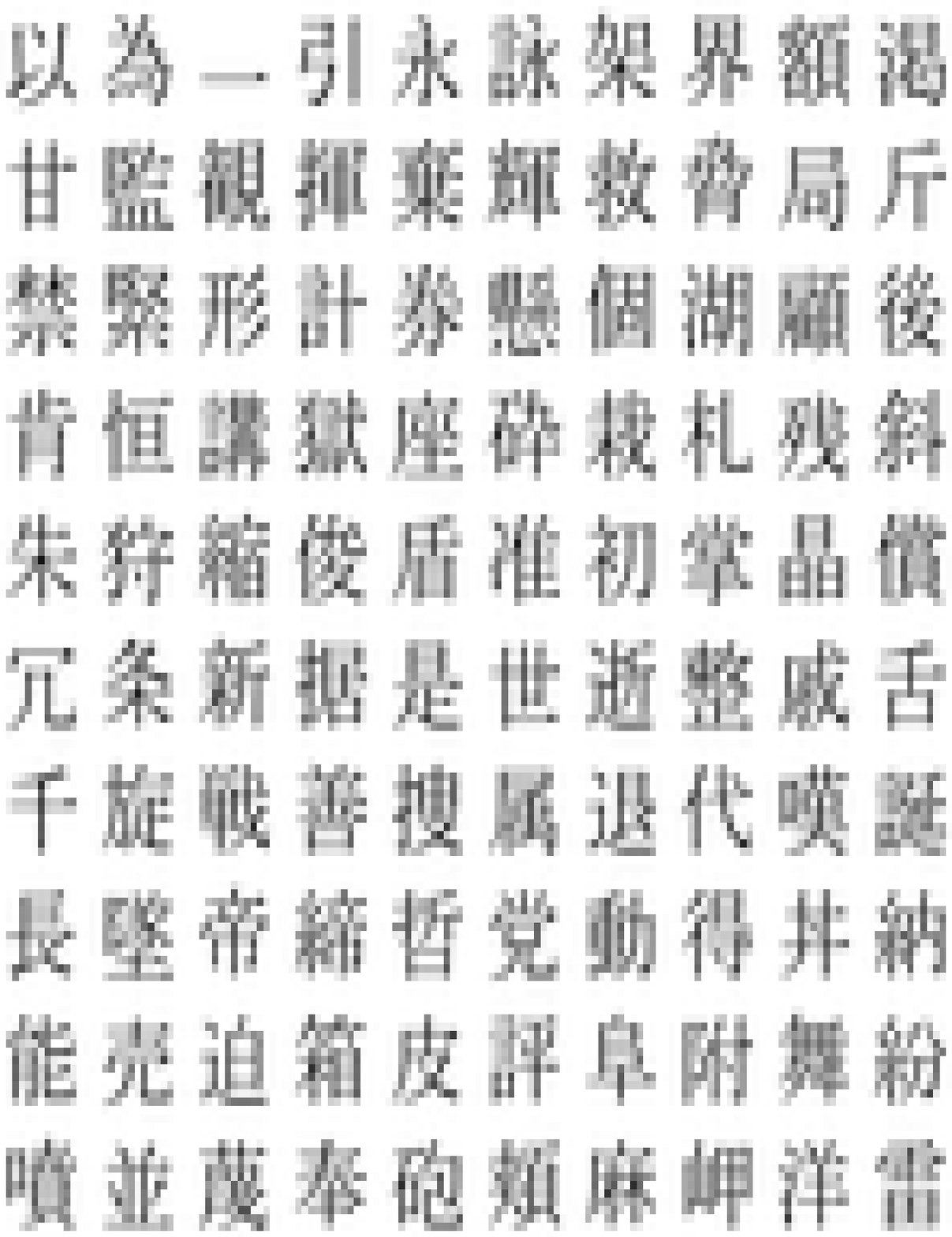} &
\includegraphics[width=0.45\textwidth]{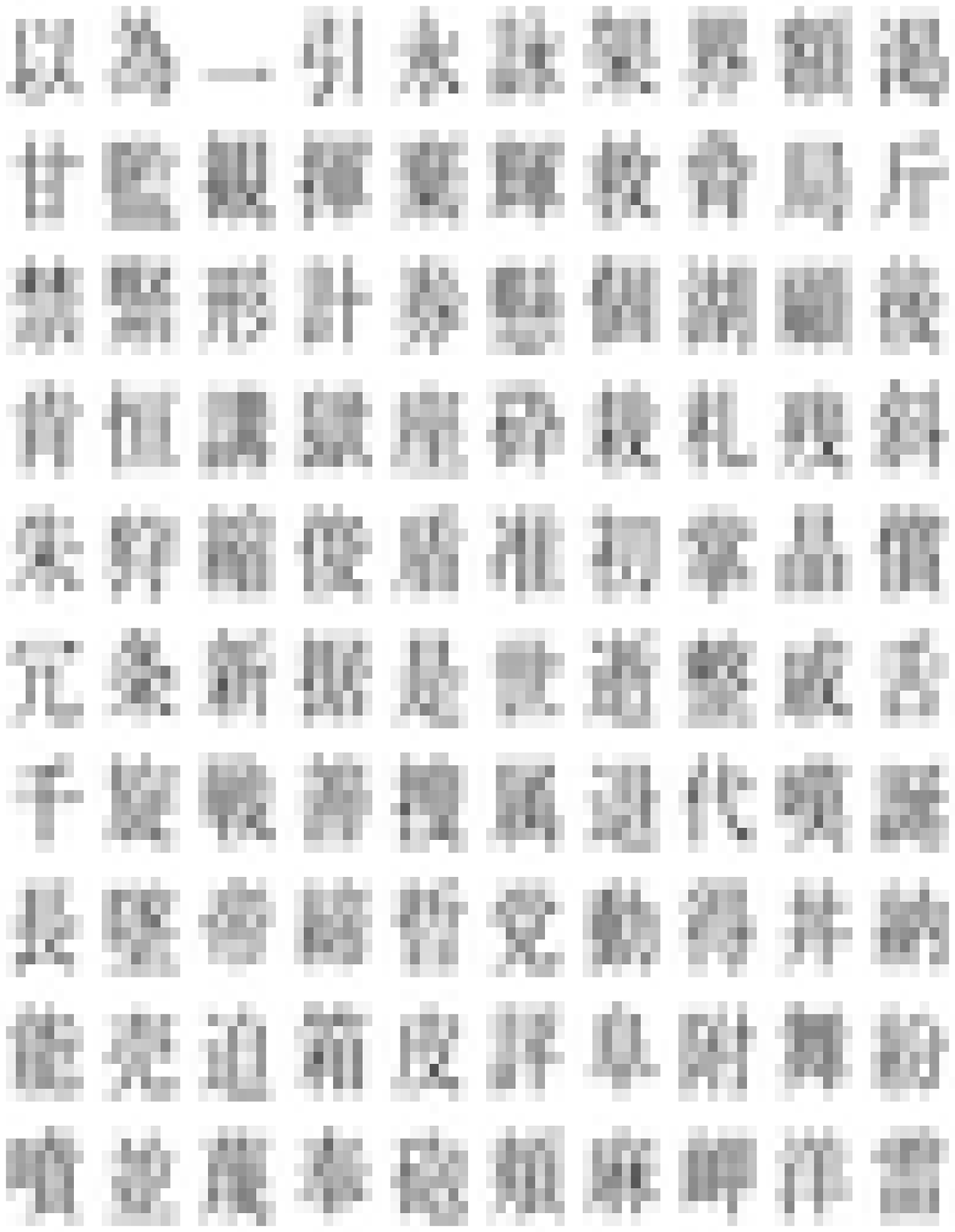} \\
112.5 ppi & 75 ppi
\end{tabular}
\caption{All evaluation kanji dataset}
\label{fig:tile_40}
\end{figure}
\end{CJK}
\end{document}